\documentclass{article}

\usepackage[square,numbers]{natbib}
\usepackage[final]{neurips_distshift_2021}

\usepackage[utf8]{inputenc} 
\usepackage[T1]{fontenc}    
\usepackage{hyperref}       
\usepackage{url}            
\usepackage{booktabs}       
\usepackage{amsfonts}       
\usepackage{nicefrac}       
\usepackage{microtype}      
\usepackage{xcolor}         

\usepackage{algorithm}
\usepackage{multirow}
\usepackage{algpseudocode}
\usepackage{subcaption}
\usepackage{amsmath,amssymb}
\usepackage{adjustbox}
\usepackage{csquotes}

\title{Distribution Shift in Airline Customer Behavior during COVID-19}
\author{%
    Abhinav Garg \\
    University of Illinois \\
    Urbana-Champaign, IL \\
    \texttt{garg19@illinois.edu}\\
    \And
    Naman Shukla \\
    Deepair LLC \\
    Dallas, TX \\
    \texttt{naman@deepair.io} \\
    \And
    Lavanya Marla \\
    University of Illinois \\
    Urbana-Champaign, IL \\
    \texttt{lavanyam@illinois.edu} \\
    \And
    Sriram Somanchi \\
    University of Notre Dame \\
    Notre Dame, IN \\
    \texttt{somanchi.1@nd.edu}
}

\begin{document}

\maketitle

\begin{abstract}
    
  Traditional AI approaches in customized (personalized) contextual pricing applications assume that the data distribution at the time of online pricing is similar to that observed during training. However, this assumption may be violated in practice because of the dynamic nature of customer buying patterns, particularly due to unanticipated system shocks such as COVID-19. We study the changes in customer behavior for a major airline during the COVID-19 pandemic by framing it as a covariate shift and concept drift detection problem. We identify which customers changed their travel and purchase behavior and the attributes affecting that change using (i) Fast Generalized Subset Scanning and (ii) Causal Forests. In our experiments with simulated and real-world data, we present how these two techniques can be used through qualitative analysis. 
\end{abstract}

\section{Introduction}

\begin{figure}[!htb]
    \centering
    \begin{subfigure}{0.6\textwidth}
        \includegraphics[width=8.5cm, height=1.2in]{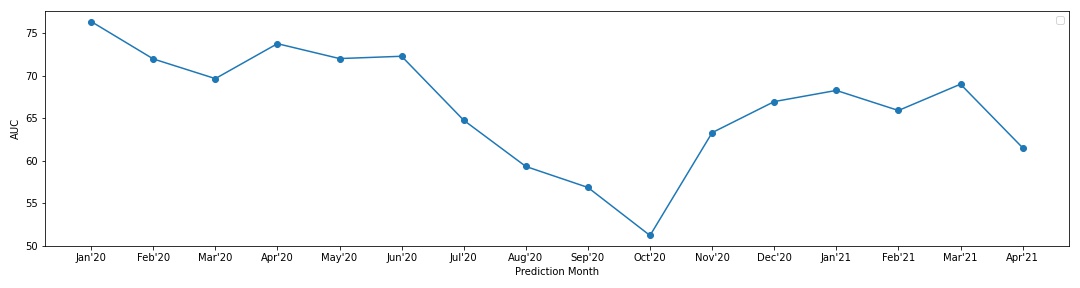}
        \caption{AUC Trend Plot}
    \end{subfigure}%
    ~ 
    \begin{subfigure}{0.4\textwidth}
        \includegraphics[width=5.5cm, height=1.2in]{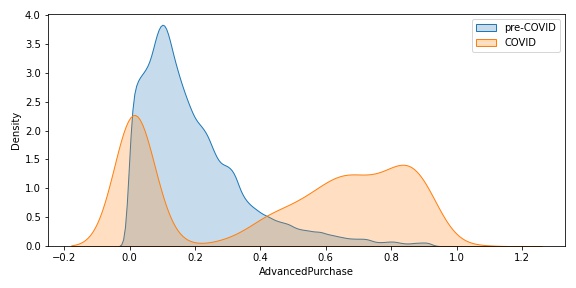}
        \caption{Density Plot}
    \end{subfigure}
    \caption{\textbf{(a)} AUC of the classifier used for purchase prediction observed from Jan 2020 to Apr 2021 (trained from Mar 2019 to Dec 2019). \textbf{(b)} Distribution of Advanced Purchase feature capturing the number of days prior to departure the customer made a booking request for pre-COVID (Mar 2019 - Feb 2020) and COVID (Mar 2020 - Sep 2020) period.}
    \label{fig:motivation}
\end{figure}

The novel coronavirus pandemic has had a seismic impact on many industries, including travel. During these unprecedented times, along with industrial operations, customer behavior also changed drastically \cite{sheth2020impact}. 
Due to this, machine learning systems built on sequential decision making were affected the most. 
Machine learning applications often implicitly or explicitly assume that data sets are drawn from stationary distributions, and the sudden shift in underlying data makes the model prone to break. Pricing based on context is one such application that is prominently driven by customers' behaviour \cite{ye2018customized, shukla2019pricing}. 
The motivation of this work stems from examining the performance of machine learning models deployed to price add-on (ancillary) products for an international airline. To illustrate, the performance of one of the deployed models that predicts the probability of ancillary purchase dropped from 75\% during training to 50\% during testing despite exhaustive offline analysis with K-fold cross validation as shown in Figure \ref{fig:motivation}. Since the onset of COVID-19, one of the features used in the model started experiencing distribution change and needed further investigation. 

Demand forecasting is frequently used for price determination given customers' context \cite{talluri2004theory, ferreira2016analytics}. Pricing applications that rely on forecasting need a distribution shift detection as the price that a customer is willing to pay can be affected by unanticipated events, such as COVID-19, that alter the system dynamics and change the data generating process. 

In this work, we (a) investigate the problem of covariate shift and concept drift for a forecasting model used for dynamic pricing, where the marginal distribution of the covariates, $P(X)$, and the conditional distibution, $P(Y|X)$, between the training and test sets differ \cite{joaquin2009datasetshift}, (b) explore two potential techniques for detecting distribution shift for online use-case, specifically, we use Fast Generalized Subset Scan \cite{daniel2013fgss} to detect covariate shift in terms of $P(X)$ and Causal Forests \cite{athey2018generalized} to identify concept drift, that is, change in $P(Y|X)$, and (c) discuss the proposal of a robust framework for contextual pricing applications in a real-world setting.  

\section{Methods}
In this section, we define how Fast Generalized Subset Scan \cite{daniel2013fgss} and Causal Forests \cite{athey2018generalized} can be used for detecting distribution shift due to shocks, such as COVID-19, experienced by a system.  

\subsection{Fast Generalized Subset Scan}
Fast Generalized Subset Scan (FGSS) is an unsupervised anomalous pattern detection technique, proposed by \citet{daniel2013fgss}. We use FGSS for covariate shift detection (change in $P(X)$) with the objective of finding shifted patterns in the test set. 
Given a set of observations $R_1...R_N$ and features $A_1...A_M$ in a test set, under the null hypothesis that there is no anomalous pattern in the test set, we use FGSS to find a subset $S^*=R^*\times{A^*}$ of self-similar groups that are anomalous, where $R^*\subseteq\{R_1...R_N\}$ and $A^*\subseteq\{A_1...A_M\}$, using a scoring function $F(S)$ defining the anomalousness of the subset $S$. If the null hypothesis is true, then the test set is generated from the same distribution as the training set. Otherwise, the training and test distributions are different with a significantly higher score $F(S)$ for those subset of observations in the test set.
\begin{equation}
     F(S) = \underset{\alpha}{\textrm{max}}F_{\alpha}(S) = \underset{\alpha}{\textrm{max}}\phi(\alpha, N_{\alpha}(S), N(S))  
\end{equation}
where $N(S)$ represents the size of the subset $S$ and $N_{\alpha}(S)$ represents the total number of p-values (obtained by passing the observed values in the test set through the inverse eCDF of the training set) that are significant at level $\alpha$ in $S$. To efficiently find the subset $S^* = arg\,max\ F(S)$, we need the function $\phi(\alpha, N_{\alpha}(S), N(S))$ to be monotonically increasing w.r.t. $N_{\alpha}$, monotonically decreasing w.r.t. $N$ and $\alpha$, and be convex \cite{daniel2013fgss}. We use the Berk-Jones statistic \cite{bj1979statistic} as $\phi$ in our experiments which satisfies these properties.

\subsection{Causal Forests}
Causal forests in a supervised method from Generalized Random Forests \cite{athey2018generalized} that estimates heterogeneity in treatment effects. A treatment effect refers to a causal effect of a treatment or intervention on an outcome variable. Causal forests can be used to estimate the Conditional Average Treatment Effect (CATE). This is useful in identifying the observations for which the treatment is positive and that benefit the most from a treatment; essentially an estimation of optimal policy assignment \cite{athey2019economists}. CATE cannot be directly observed for a unit because of the “fundamental problem of causal inference” \cite{holland1986causal}, making it impossible to observe unit-level causal effects and the reason why we can never directly observe the counterfactual condition of a unit of observation. For each observation $X \in \mathbb{R}^{m}$ where $m$ is the number of covariates, there are two potential outcomes $Y_1$ and $Y_0$ corresponding to the binary treatment variable $D\in \{0,1\}$, but only one of them is observed. The conditional expectation of an outcome for the treatment or control, $\mu_{d}(x)$ is defined as:
\begin{equation}
    \mu_{d}(x) = E[Y_{i}|X_i=x, D_i=d] \quad with \quad D \in \{0,1\}
\end{equation}
and CATE ($\tau(x)$) is the difference in expectation of the potential outcomes given x,
\begin{equation}
    \tau(x) = E[Y_{i}^1-Y_{i}^0|X_i=x] = \mu_{1}(x) - \mu_{0}(x) \label{eq: tau}
\end{equation}
We use causal forests to estimate the causal effect of COVID-19 intervention and identify the concept drift $P(Y|X)$ in the system. This requires some data observed post-intervention to be used for training so as to learn the unit-level interventional change in treatment.

\section{Experiments}
Ancillaries are optional products or services sold by businesses to complement their primary product \cite{bockelie2017incorporating}. In this work, we utilized the following datasets from the airline industry for an ancillary market: (1) simulated interaction of ancillary pricing, and (2) real-world ancillary booking requests data from a large airline containing price variability.

\subsection{Simulated dataset}
To test our approach, we generated a simulated dataset of customer and flight seat (ancillary) interactions using open-sourced flight simulator \footnote{https://github.com/deepair-io/flai}. 
In this dataset, we artificially varied the arrival rates of the customers in train and test data. 
The details of the data-generation process can be found in Appendix \ref{ref:exp_setup}. Figure \ref{fig:sim}(a) shows the generated data distribution. The training set represent the control period ($d=0$) and test set represents the treatment period ($d=1$). Figures \ref{fig:sim}(b) and (c) show the results from Causal Forests and FGSS, respectively. FGSS is able to identify 26.01\% points in the test set as shifted. The shifted points have \textit{AdvancedPurchase} feature values greater than 5000 despite the test distribution having a peak around 4000. This is because the train distribution has few points in the values between 0 and 4000, which the FGSS model identifies as normal behavior. On the other hand, Causal Forests is able to identify 23.26\% points with positive shift and only 0.56\% points with negative shift, indicating a superior performance of Causal Forests model to identify shift on intervention. The positive shift occurs over the range of test set values, which aligns with the distribution of test set simulated significantly different from the train set. 
 
\begin{figure}[!htb]
    \centering
    \begin{subfigure}{0.33\textwidth}
        \centering
        \includegraphics[width=5.5cm, height=1.2in]{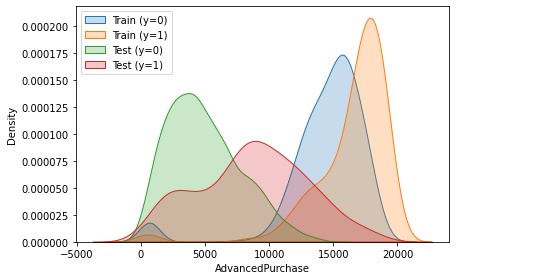}
        \caption{Data Distribution}
    \end{subfigure}%
    ~
    \begin{subfigure}{0.33\textwidth}
        \centering
        \includegraphics[width=5.5cm, height=1.2in]{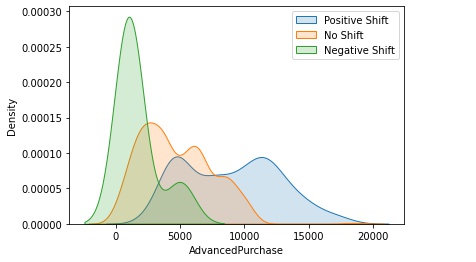}
        \caption{Causal Forests}
    \end{subfigure}%
    ~ 
    \begin{subfigure}{0.33\textwidth}
        \centering
        \includegraphics[width=5.5cm, height=1.2in]{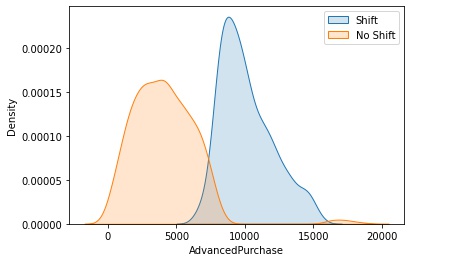}
        \caption{FGSS}
    \end{subfigure}
    \caption{\textbf{(a)} Train and test distributions for the binary target ($y$) of purchase/no-purchase. \textbf{(b)} Density plot of customers identified having a positive, negative and not shifted by Causal Forests in the test set. Following Equation \eqref{eq: tau}, \textit{positive shift} identified by Causal Forests represent the customers that are more likely to purchase an ancillary in the treatment period, ($P_{1}(Y|X=x_i) > P_{0}(Y|X=x_i)$ with significance $\alpha$) and similarly \textit{negative shift} represent the customers that are more likely to purchase in the control period, ($P_{1}(Y|X=x_i) < P_{0}(Y|X=x_i)$ with significance $\alpha$), while \textit{no shift} represent the customers whose probability of purchase in the train and test sets are similar ($P_{1}(Y|X=x_i) \approx P_{0}(Y|X=x_i)$). \textbf{(c)} Density plot of customers identified as shifted ($P_1(X=x_i) \neq P_0(X=x_i)$ with significance $\alpha$) and not shifted ($P_1(X=x_i) \approx P_0(X=x_i)$ with significance $\alpha$) by FGSS in the test set. FGSS only tells whether an observation has a covariate shift or not, and says nothing about the concept drift.}
    \label{fig:sim}
\end{figure} 

\subsection{Real-world dataset} \label{real-world}
The real-world dataset consists of customer booking requests from March 2019 to September 2020. We use this dataset to evaluate the performance of FGSS and Causal Forests in detecting distribution shift due to COVID-19 in real-world setting. The airline identifies the time period starting March 2020 as the “COVID” era because a large portion of scheduled flights started getting canceled/rescheduled and observed decreased ticket sales. We use five features in our experiments - \textit{AdvancedPurchase}, \textit{LengthOfStay}, \textit{GroupSize}, \textit{TotalDuration} and \textit{TripType}. The description of the features can be found in Appendix: Table \ref{tab:des}. 
Figure \ref{fig:cf} and \ref{fig:fgss} show the results obtained on the hold-out test set. Both Causal Forests and FGSS results indicate that there is significant shift in all of the features. \textit{LengthOfStay} feature has shift towards value $0$ (indefinite stay at destination), indicating most customers are not opting for vacation/business travel but essential movement only. The same conclusion can be drawn from the shift observed in \textit{TripType} feature with customers preferring to travel one-way. \textit{GroupSize} feature has a shift towards value $1$, indicating customers are traveling alone and not in large groups during the pandemic. \textit{AdvancedPurchase} has a shift towards smaller values indicating customers are not booking tickets way ahead of the travel date due to the uncertain nature of the pandemic. \textit{TotalDuration} has a shift towards smaller values as well, indicating the operational changes made by the airline to operate on shorter routes during the pandemic.

\begin{figure}[!htb]
\centering
  \includegraphics[width=14cm, height=1.2in]{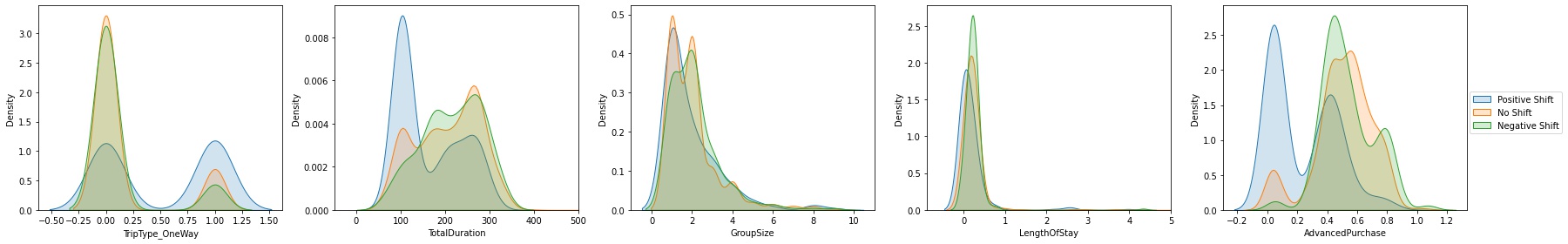}
\caption{Density plot of observations identified as positive, negative and not shifted by Causal Forests.}
\label{fig:cf}
\end{figure}

\begin{figure}[!htb]
\centering
  \includegraphics[width=14cm, height=1.2in]{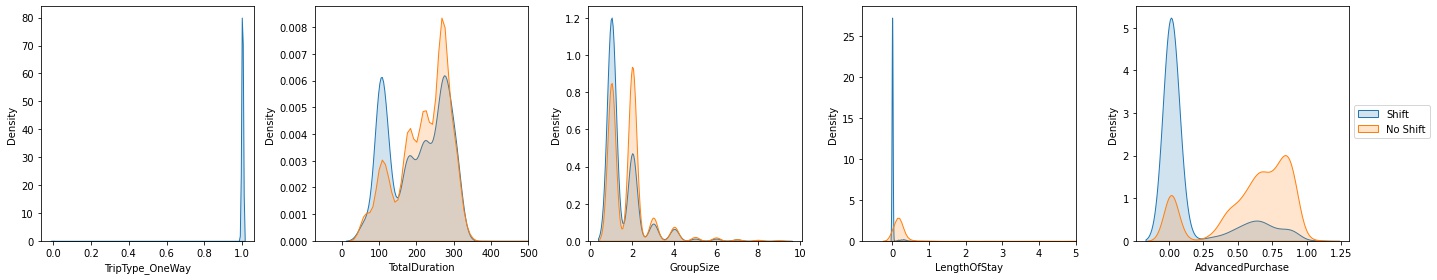}
\caption{Density plot of observations identified as shifted and not shifted by FGSS.}
\label{fig:fgss}
\end{figure}

\section{Conclusion}
We explored two techniques, Fast Generalized Subset Scan and Causal Forests, for covariate shift detection when a system experiences a shock. We applied these techniques to an airline ancillary purchase use case and saw that some of the features used for ancillary pricing experienced a covariate shift while the probability of purchase experienced a concept drift during COVID-19. We identified the observations in the test set that follow a significantly different distribution compared to the training set. Causal forests, while able to detect a concept drift in real-time, need a significant amount of data post-intervention for training. On the other hand, FGSS can only detect covariate shift on a batch of data but does not need data post-intervention for training. Hence, there is a possibility to combine the two approaches for a unified and robust distribution shift detection. At the same time, the construction of a good testing pipeline, along with an agreement by multiple models as well as a visual inspection, can jointly help identify shift patterns. These patterns can then be cross-checked with designers' hypothesis and domain experts' knowledge for validation and potential changes to the system before being corrected for enhanced model performance.  In the future, we aim to extend this work to other applications, test multiple shift correction approaches and provide recommendations for adapting to a shift induced by sudden shocks to the system (see Appendix \ref{ref:pricing_model} for details).

\begin{ack}
We sincerely and gratefully acknowledge our airline partners for their continuing support. The academic partners are also thankful to deepair (www.deepair.io) for funding this research.
\end{ack}

\bibliographystyle{plainnat}
\bibliography{main}


\newpage

\appendix

\section{Appendix}

\subsection{Additional Information} \label{ref:exp_setup}
 The simulated dataset consists of interactions for selecting flight seat after the ticket is purchased. Customer arrivals are simulated using non-homogeneous poisson process \cite{lewis1979simulation}. Multinomial logit is used for customer choice model in the simulator \cite{hausman1984specification}. The offered prices are randomly sampled from minimum to maximum allowed price for that flight. The intervention is artificially created by changing the customer arrival rate. We have simulated 2 datasets with different parameters of the poisson process for 10 long-distance flights as train and test (Fig. \ref{fig:sim}(a)). We use the arrival rate (\textit{AdvancedPurchase}) and number of seats sold (\textit{Sold}) as features, with \textit{AdvancedPurchase} feature having an apparent covariate shift. For Causal Forests, we leak 40\% points randomly sampled from the test set into training.

Table \ref{tab:des} presents the description of features discussed in Section \ref{real-world}. FGSS requires all features to be categorical. We discretize all numeric features by binning them at quartiles for FGSS training. We divide the real-world data into two parts: pre-COVID (March 2019 to February 2020) and COVID (March 2020 to September 2020). We use the pre-COVID period for training FGSS and Causal Forests model. Further, we leak March 2020 to August 2020 COVID data into training for Causal Forests, and leave September 2020 as the hold-out test set to report results.

\begin{table}[ht]
\centering
\caption{Feature Description}
\begin{adjustbox}{max width=1.0\textwidth,center}
\begin{tabular}{ccc}
\toprule
Feature & Type & Description \\
\hline
AdvancedPurchase & Numeric & Number of days prior to flight departure the request has been made \\
LengthOfStay & Numeric & Number of days between onward and return flight (zero for one-way ticket) \\
GroupSize & Numeric & Number of passengers on the customer booking \\
TotalDuration & Numeric & Duration of the flight in minutes \\
TripType & Categorical & Indicates whether the customer has booked "One-Way" or "Round-Trip" \\
\hline
\end{tabular}
\end{adjustbox}
\label{tab:des}
\end{table} 

\subsection{Robust Pricing Model} \label{ref:pricing_model}
\begin{figure}[!htb]
\centering
  \includegraphics[width=10.5cm, height=3.5cm]{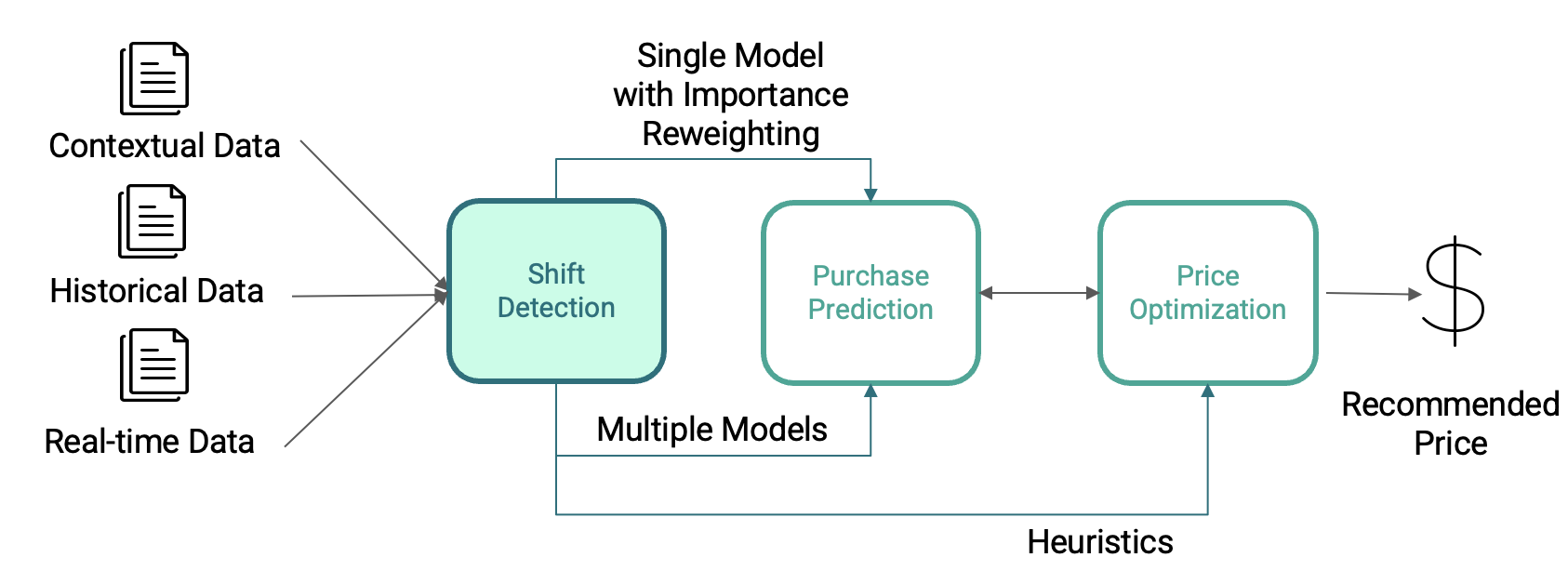}
\caption{Proposed pricing framework.}
\label{fig:pricing}
\end{figure}

Shukla et al. \cite{shukla2019pricing} proposed a two-stage pricing model that uses the purchase probability prediction for price recommendation. Figure \ref{fig:pricing} shows the proposed pricing framework with shift detection before the supervised model prediction. The shift detection layer first checks if there is a shift in the system or not. Causal Forests being a supervised technique can be used to detect shift "on-line". However, Causal Forests require data observed after intervention for training and identification of the start a shift induced by an intervention to the system is usually a difficult task. We propose to take advantage of the unsupervised nature of FGSS to achieve that. FGSS can learn the normal behavior from historical data and run in batch mode to detect anomalous patterns during testing. If the percentage of observed anomalous patterns breach a user-defined threshold for a significant period of time, Causal Forests can use that data as treatment for training.

Once the shift has been detected in "real-time", it can be handled in multiple ways: a single model trained after reweighting shifted points, multiple models - one for each shifted pattern, or heuristics based on domain knowledge to better price the product. We aim to perform further experiments in future to test the feasibility of these 3 approaches and propose an efficient strategy to handle covariate shift for decision-making under uncertainty.

\end{document}